\documentclass[11pt]{article}
\usepackage{umd_scholarly_paper}
\usepackage{times}
\usepackage{url}
\usepackage{latexsym}
\usepackage{graphicx}
\usepackage{caption}

\title{A Review on Semi-Supervised Relation Extraction}

\author{Yusen Lin \\
  University of Maryland \\
  {\tt yusenlin@umd.edu} \\}

\date{}

\begin{document}
\maketitle
\begin{abstract}

Relation extraction (RE) plays an important role in extracting knowledge from unstructured text but requires a large amount of labeled corpus. To reduce the expensive annotation efforts, semi-supervised learning aims to leverage both labeled and unlabeled data. In this paper, we review and compare three typical methods in semi-supervised RE with deep learning or meta-learning: self-ensembling, which forces consistent under perturbations but may confront insufficient supervision; self-training, which iteratively generates pseudo labels and retrain itself with the enlarged labeled set; dual learning, which leverages a primal task and a dual task to give mutual feedback. Mean-teacher \cite{tarvainen2017mean}, LST \cite{li2019learning}, and DualRE \cite{lin2019learning} are elaborated as the representatives to alleviate the weakness of these three methods, respectively.

\end{abstract}

\section{Introduction}

Relation extraction (RE) refers to the automatic identification of the relations given the entity mentions. For example, given "Steve Jobs is the founder of Apple", the "found\_by" relation can be extracted between the entities "Apple" and "Steve Jobs". RE is essential in many downstream applications like knowledge graph construction and question answering and inference. However, it typically requires a massive labeled corpus, especially for domain-specific RE. 

To reduce labeling effort, it is significant to develop techniques like active learning, distance supervision, semi-supervised learning, unsupervised learning, in-context learning, and etc. Active learning reduces the labeling cost by selecting the most uncertain samples for the annotator to annotate, but still requires human-in-the-loop effort; distance supervision exploits an external knowledge base; semi-supervised learning leverages both labeled and unlabeled data; unsupervised learning relies on only unlabeled data but so far needs further improvement; in-context learning in GPT-3 \cite{floridi2020gpt} learns the knowledge during inference, but more progress in research is still needed. This paper focuses on applying semi-supervised learning on RE tasks through deep learning and meta-learning. 

\begin{figure*}[ht]
\centering
\includegraphics[width=\textwidth]{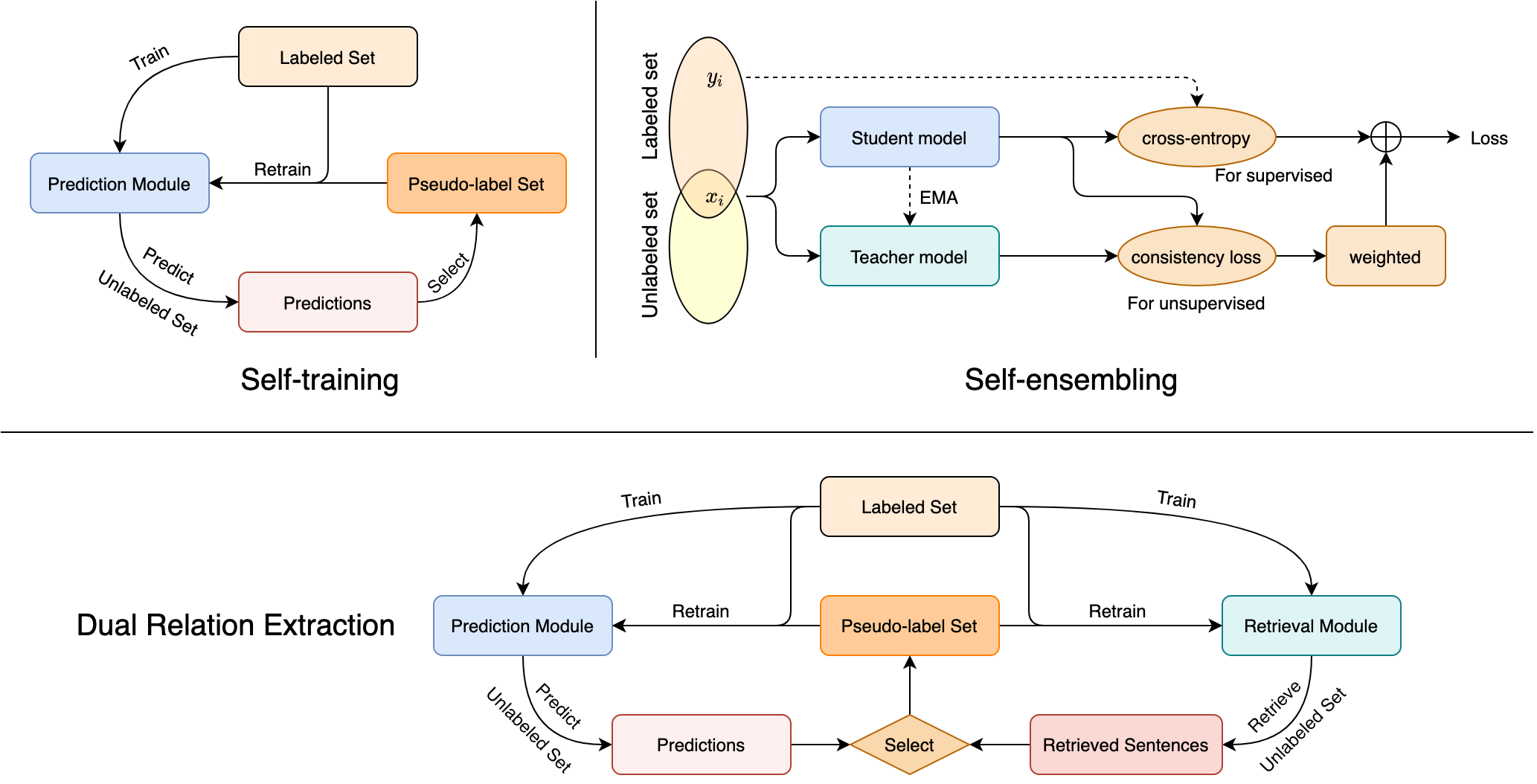}
\caption{\textbf{Difference between self-training, self-ensembling and dual learning.} }
\label{fig:diff_semi_methods}
\end{figure*}

Three recent approaches, including self-ensembling, self-training, and dual training, are mainly covered.
Their principles and differences are shown in figure \ref{fig:diff_semi_methods}, where mean-teacher \cite{tarvainen2017mean} and DualRE \cite{lin2019learning} are chosen as the examples for self-ensembling and dual learning, respectively.
Self-ensembling assumes the model remains consistent under perturbations on the model parameters or the input data. However, with limited labeled data, it is likely to encounter insufficient supervision because no additional training data is introduced. In contrast, self-training predicts pseudo labels and retrains the model iteratively with the continuously enlarged labeled set. Despite avoiding insufficient supervision, it inevitably suffers from a gradual drift due to biased predictions and accumulated errors. To alleviate both issues, dual learning forms a close loop between a primal task and a dual task to obtain mutual feedback signals and boost the entire learning process. For RE, the duality occurs on retrieving mentions given a specific relation and predicting relations given a specific mention. 

This paper is organized as follows: Section 2 introduces self-ensembling and its three advanced approaches; Section 3 talks about the pros and cons of self-training and one recent way, LST \cite{li2019learning}, to overcome the gradual drift by meta-transfer-learning; Section 4 discusses the difference between dual learning in RE and the above-mentioned methods, followed by the description of DualRE \cite{lin2019learning}; Finally, Section 5 presents our conclusions and future research directions.

\section{Self-ensembling}

Self-ensembling comes from the general idea that better performance can be yielded by ensembling multiple models. It extends the idea by ensembling outputs from models under various regularization and input augmentation conditions or on different training epochs. The concept behind it is to assume no changes in the prediction of the unlabeled data even if small perturbations exist on the model parameters or on the data. 

In this article, we describe three advanced implementations of self-ensembling, including $\Pi$-model, temporal ensembling \cite{laine2016temporal}, and mean teachers \cite{tarvainen2017mean}.

\subsection{$\Pi$-Model}

Figure \ref{fig:temporal_emsembling_structure} demonstrates the idea of $\Pi$-model: given the same input with various dropout conditions, $\Pi$-model imposes consistent predictions between two realizations. 
Technically, for each training sample $x_i$ , where $i$ stands for the sample index, the network would predict twice, leading to output vectors $z_i$ and $\tilde{z_i}$ . 
The loss function is composed of supervised and unsupervised loss terms. The former is a standard cross-entropy loss, only evaluated on labeled data, and $z_i$ is fed to it; the latter takes the mean square loss between $z_i$ and $\tilde{z_i}$ and it is evaluated on all data including the unlabeled ones. 
Besides, the mean square loss is multiplied by a time-varying weighting function $w(t)$. 
The weighting function $w(t)$ starts from 0, and then slowly ramps up along the Gaussian curve after training a number of epochs. 
The final loss function is the sum of these two loss terms.

Compared to traditional self-ensembling methods, which only force the final classification to be consistent, it directly penalizes the network by comparing the difference between two output vectors $z_i$ and $\tilde{z_i}$ given the same input $x_i$ , imposing a much stronger constraint. However, the training curve is readily jitter since the $\tilde{z_i}$ is calculated based on a single training step. Meanwhile, it would consume 2x training time due to evaluation twice per step.

\subsection{Temporal Ensembling}

\begin{figure*}[ht]
\centering
\includegraphics[width=\textwidth]{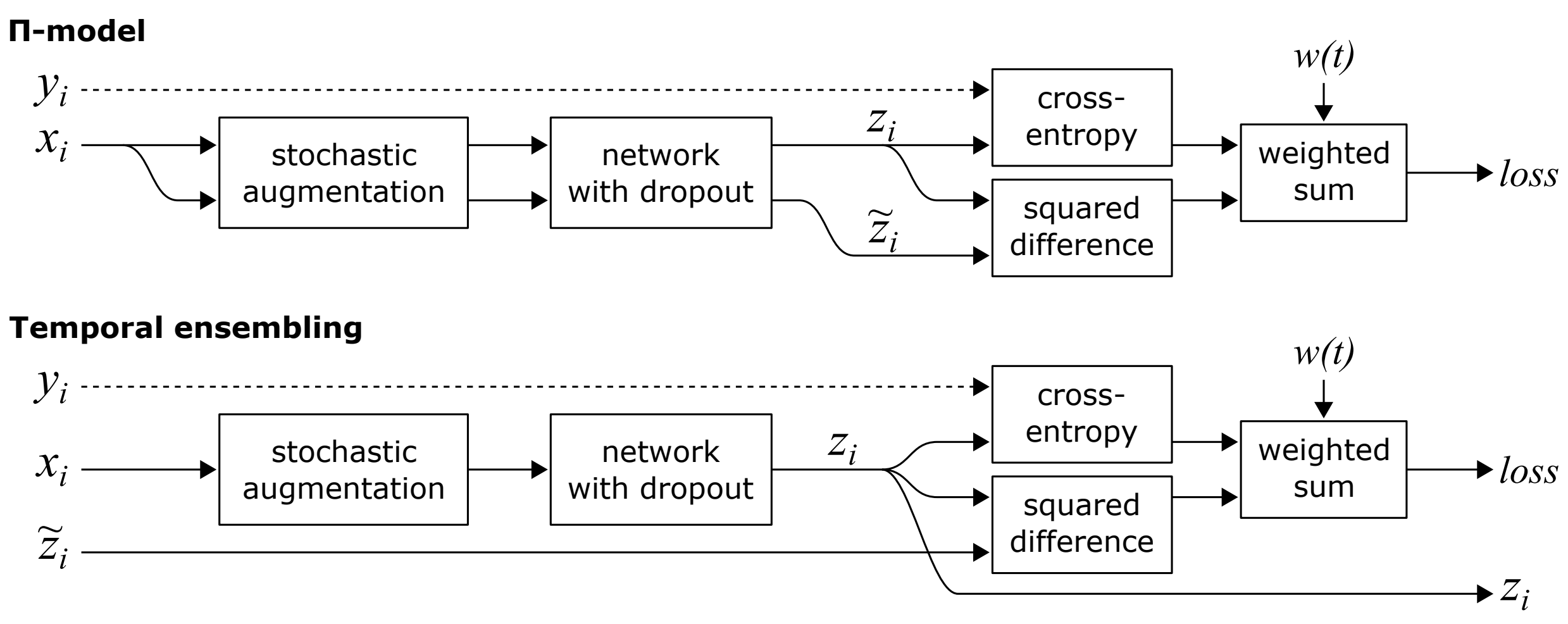}
\caption{\textbf{Difference between $\Pi$-model and Temporal ensembling} \cite{laine2016temporal}.}
\label{fig:temporal_emsembling_structure}
\end{figure*}

The difference between $\Pi$-model and temporal ensembling is shown in figure \ref{fig:temporal_emsembling_structure}.
To overcome the weakness of $\Pi$-model, temporal ensembling leverages the historical predictions. Instead of evaluating twice, it maintains an exponential moving average (EMA) for the previous prediction as $\tilde{z_i}$, reducing the noise and avoiding inference twice.

Due to dropout regularization and model learning, the output vector $z_i$ varies every training epoch. 
To aggregate these historical predictions with a larger weight lying on the recent epochs, a matrix Z is provided and is updated through a momentum method. 
Every training epoch, the output vector $z_i$ is stored into $Z$ by $Z_i = \alpha Z_i + (1-\alpha) z_i$. 
The $\alpha$ is the momentum term that 5 determines the weights of the recent information. 
Compared with $\Pi$-model, the target vector $\tilde{z_i}$ is replaced with the $Z_i$ in this method. 
However, the $Z_i$ needs to divide by a factor $1 - \alpha^t$ for correction of the startup bias before being assigned to $\tilde{z_i}$, where t is the epoch index.

With the temporal design of the target vectors $\tilde{z_i}$, its training is much more efficient than the training of $\Pi$-model.
Most importantly, the temporal accumulation of Z would make $\tilde{z_i}$ more consistent, less jitter, and less noisy, compared to being generated every step in $\Pi$-model, which leads to better performance in the empirical experiments of \cite{laine2016temporal}. 
As for the weakness, it may requires a relatively large memory due to the matrix Z, especially if given a large dataset.

\subsection{Mean Teachers}

Compared to the EMA on historical prediction in temporal ensembling, mean teacher, whose principle is shown in the self-ensembling part of figure \ref{fig:diff_semi_methods}, applies the EMA on model weights. 
It fastens the pace of aggregating learned information, instead of only updating the target vector $\tilde{z_i}$ once per epoch. 
Without the limits of the matrix $Z$ in temporal ensembling, it satisfies a larger dataset and works well with online learning. 
Moreover, the experiments in \cite{tarvainen2017mean} prove that it could produce better performance and require less training time.

Mean teacher is composed of a student and a teacher model. 
The student model shares a common role with the models of $\Pi$-model and temporal ensembling, learned through a classification loss with labeled data and a consistency loss with all data. 
One core difference is that the target vector $\tilde{z_i}$ is predicted by the teacher model instead of the student one. 
Since \cite{polyak1992acceleration} proposed averaging the model parameters over the training steps tends to predict more accurately, the teacher model shares the same structure with the student one and simply takes the EMA of the student model weights after every training step, without any learning.

\section{Self-training}

Self-training is the process of iteratively retraining the model with labeled and pseudo-label data for improvement. The pseudo-label set is enlarged continuously by high-confidence predictions of the model over the training. It solves the insufficient-supervision limits of self-ensembling, resulting in, usually, its outperforming self-ensembling in empirical studies, e.g., \cite{li2019learning}, especially in the case with scarce labeled data. However, they may work in a complementary way.

Since the pseudo labels come from the model predictions, which could be biased, most existing self-training methods inevitably bring the noise to the labeled set. These cumulated errors may finally hurt the model performance, which is called 'semantic drift’. To alleviate semantic drift, a recently proposed method, LST \cite{li2019learning}, leverages meta-transfer-learning \cite{sun2019meta} to cherry-pick the pseudo labels meanwhile fine-tunes the model with only the labeled set after each training step.

\begin{figure*}[ht]
\centering
\includegraphics[width=\textwidth]{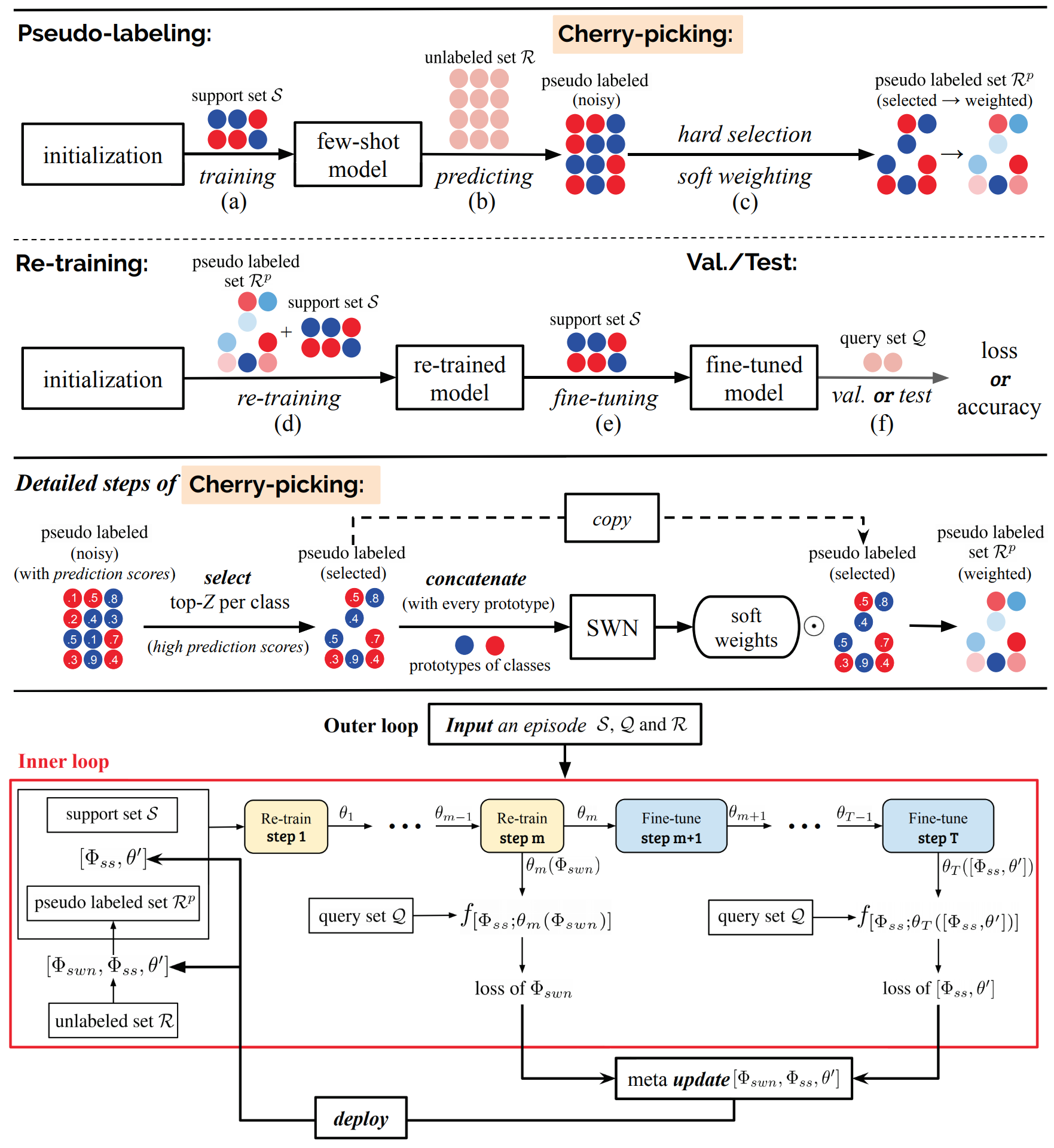}
\caption{\textbf{The computing flow of LST} \cite{li2019learning}.}
\label{fig:LST}
\end{figure*}

\subsection{Learning to Self-train (LST)}

The computing flow of LST can be seen in figure \ref{fig:LST}, which consists of pseudo labeling, cherry-picking, iterative retraining and fine-tuning model, and meta-optimizing. 
The model is first pre-trained on a support set $S$ and then pseudo-labels the unlabeled samples. 
Following cherry-picking the pseudo labels by hard selection and soft weighting, the model is retrained on the support set $S$ and the picked labeled set $R^p$ and then fine-tuned on only the support set $S$. 
The final step is testing on a query set $Q$, acting as a validation loss for the meta-learning optimization process of LST.

\subsubsection{Pseudo-labeling}

Like every common self-training method, the first step is to pre-train a classifier $\theta$, and then predict the pseudo labels $Y^R$ for the unlabeled set $R$ with the classifier $\theta$.

Theoretically, any model is applicable for $\theta$. Meta-transfer-learning \cite{sun2019meta} is applied in \cite{li2019learning}, which includes the inner loop and the outer loop. 
The former is to train $\theta$ for a specific task with $S$ , while the latter is to scale and shift the initialization weights of the pre-trained network $\theta'$ by meta-learning for quick fine-tuning. 
$R$ is fed into $\theta$ for getting $Y^R$ after fine-tuning.

\subsubsection{Cherry-picking}

Cherry-picking is effective for reducing label noises and improving performance, whose steps include the hard selection and soft weighting network (SWN). Hard selection is to pick up top $Z$ samples per class based on the confidence scores of $Y^R$. Given $C$ classes, there are $ZC$ samples in the “cherries” $R^p$. The "cherries" $R^p$ is then reweighted by a meta-learned SWN before feeding to self-training. SWN tends to increase the weights of better pseudo labels and decrease the worse ones. It first computes the prototype feature of each class, which averages the features of the corresponding class on $S$. Then, the sample feature is concatenated with it as the input of SWN. Consequently, the weights w, output of SWN, somehow reflect the distance between the pseudo labels and the prototype representations of $C$ classes.

\subsubsection{Iterative retraining and fine-tuning}

Supposed $T$ steps exist in the inner loop, $\theta$ is first trained on both $S$ and $R^p$ for $m$ steps ($\theta_m$), 
then is fine-tuned on only $S$ for $T - m$ steps ($\theta_T$). 
Before predicting $Y^R$ with the fine-tuned $\theta_T$, the unlabeled set $R$ is split into multiple subsets, each retraining occurs on a new subset $R'$. 
$\theta_T$ then predicts on $R'$ to obtain more accurate pseudo labels $Y^R$ and enlarges $R^p$ before the recursive self-training.

\subsubsection{Meta optimizing}

The inner loop is for learning $\theta$ in few steps with machine learning or deep learning while the outer loop meta-optimizes for the ability to fast adapt a pre-trained network $\theta'$ to a new specific task. 
There are multiple parameters to be meta-learned, including $\phi_{SWN}$ parameters in SWN and $\phi_{\theta}$ parameters in $\theta$. 
The impact of $\phi_{SWN}$ only reflects on retraining through the generated soft weights while $\phi_{\theta}$ affects the entire self-training process. 
To better serve the effect of the parameters, $\phi_{SWN}$ and $\phi_{\theta}$ are updated in different self-training stages. The losses of the retraining $\theta_m$ and the fine-tuned $\theta_T$ are applied for meta-optimizing $\phi_{SWN}$ and $\phi_{\theta}$ , respectively, with different meta-learning rates.

\section{Dual learning}

As mentioned in sections II and III, self-ensembling suffers from insufficient supervision due to no extra pseudo labels generated, which limits the model improvement, while biased predictions and accumulated errors inevitably make self-training subject to the gradual drift. To alleviate these issues and enhance the learning process, Dual learning leverages two related but complementary tasks to optimize and regularize themselves. Specifically, for relation extraction, recently, one of the state-of-the-art method, DualRE \cite{lin2019learning}, proposes the duality between retrieving sentences given a certain relation and predicting the relation label given a specific sentence.

\subsection{Dual Relation Extraction (DualRE)}

Different from self-training that one model is trained to provide additional pseudo-label data, one primal and dual modules, the prediction and retrieval modules, exist in DualRE and are jointly optimized to enrich the training set and enhance each other. The prediction module $P$ aims for relation extraction by predicting relation labels given a sentence and aims to evaluate the results from the retrieval module. The retrieval module $R$ ranks sentences based on relevant scores given a certain relation label and outputs the top ones. The retrieved sentences from the unlabeled corpus are then served as pseudo-label data for the prediction module. 
The concept and the interaction between two modules are shown in the dual learning of figure \ref{fig:diff_semi_methods}.
Since both modules provide pseudo labels, insufficient supervision is solved, while two modules regulate each other during the joint learning process, leading to less biased pseudo labels and alleviation of semantic drift. As for the objective function, it consists of three terms, $O = O_P + O_R+O_U$, where $O_P$, $O_R$, and $O_U$ are for the prediction module $P$, retrieval module $R$, and unsupervised part, respectively.

\subsubsection{Relation prediction module}

Prediction module $P_{\theta}$, where $\theta$ is the parameters, mainly works for predicting the relations given the mentions. 
An encoder is applied to encode the mention pair and the corresponding context into a representation vector $z$. 
Vector $z$ is then fed into a simple softmax classifier for relation classification. For the encoder, it could be any feature extractor, e.g., RNN-based, CNN-based, or Transformer-based. 
For the objective function, $O_P = E_{(x,y)\in L}[logP_{\theta}(y|x)]$, cross-entropy is used given $x$ as the mention pair and its context, $y$ as the relation label, and $L$ as the labeled set.

\subsubsection{Sentence retrieval module}

Retrieval module $Q_{\phi}$ is for retrieving the most relevant mentions $x$ given a certain relation $y$, where $\phi$ is the parameters. In \cite{lin2019learning}, Learning-to-rank \cite{liu2009learning} is used to rank the relevance scores of the mentions so that top mentions can be obtained. 
Learning-to-rank intends to learn the joint distribution $Q_{\phi}(x,y)$ and its corresponding objective function is $O_R = E_{(x,y)\in L}[logQ_{\phi}(x,y)]$.
Since learning joint distribution $Q_{\phi}(x,y)$ is intractable due to the high computation cost of traversing all possible pairs of $(x,y)$ and $Q_{\phi}(x,y)$ is in proportion to $Q_{\phi}(x,y)$ when given fixed probability of relation label $Q(y)$, \cite{lin2019learning} learns $Q_{\phi}(x,y)$, instead. Pointwise and pairwise approaches are available for learning-to-rank and empirical experiments in \cite{lin2019learning} demonstrate the former one generally performs better.

\textbf{Pointwise approach:} $Q_{\phi}(x,y)$ outputs the relevance scores between the set of mentions ${x}$ and the relation label y for the retrieval module. 
By taking the inner product between the encoded representation z from x and the relation embedding $y_e$ from $y$ , followed by a sigmoid function, $Q_{\phi}(x|y) = \sigma(z^Ty_e)$, it forces $Q_{\phi}(x|y)$ close to 1 when $(x, y)$ are in the labeled set and close to 0 when $(x,y')$ are not the correct pairs. 
$E_{(x,y)\in L}[log\sigma(z^Ty_e)]+E_{(x,y')\in L}[log(1-\sigma(z^Ty_e'))]$ is the specific objective function, where the former term maximizes the probabilities of the correct pairs whereas the latter minimizes the probabilities of the incorrect pairs.

\subsubsection{Interaction in the EM-based joint optimization process}

Different from self-training, DualRE generates pseudo labels under the cooperation of two modules. 
This collaboration helps reduce biased predictions by regularizing the outputs of each other through majority voting. 
Specifically, it takes the intersection instances as the final pseudo labels. This process effectively alleviates both the insufficient supervision and the semantic drift and improves the overall model performance. 
The objective function of the unsupervised part is $O_U=E_{x\in U}[logQ_{\phi}(x)] \geq E_{x\in U, y\sim P_{\theta}(y|x)}[log\frac{Q_{\phi}(x,y)}{P_{\theta}(y|x)}]$. It is derived from the EM algorithm \cite{dempster1977maximum} where $U$ is the unlabeled data, $P_{\theta}$ is the prediction module, and $Q_{\phi}$ is the retrieval module. 
The equal sign is available only if $P_{\theta}(y|x)=Q_{\phi}(y|x)$, which may be achieved by taking the results that two modules agree with. The overall objective function $O$ is optimized with EM algorithm, alternatively updating the two modules by maximizing the lower bound of $O_U$ (M step) and updating the prediction module $P_{\theta}$ (E step).

In the E step, the prediction module $P_{\theta}$ is updated by fixing the retrieval module $Q_{\phi}$ , which corresponds to minimizing the KL divergence $KL(P_{\theta}(y|x)\| Q_{\phi}(y|x))$. Using the wake-sleep algorithm [9], minimizing the reversed KL divergence $KL(Q_{\phi}(y|x)\| P_{\theta}(y|x))$ obtains the same result $P_{\theta}(y|x)=Q_{\phi}(y|x)$. 
For the M step, similar to the E step, the retrieval module $Q_{\phi}$ is updated by fixing the prediction module $P_{\theta}$. Each E and M steps train the modules with a shared version of labeled and pseudo-label data to boost the overall model performance.

\section{Conclusions and future research directions}

Relation extraction is essential in information extraction from unstructured text. However, it usually requires huge annotation efforts, especially for domain-specific cases. This article aims to introduce three recent methods in semi-supervised learning for RE, which include self-ensembling, self-training, and dual learning. Self-ensembling forces consistency under perturbations but may have the insufficient supervision issue; self-training recursively yields the pseudo labels and retrains the model with the enlarged labeled set but may suffer from gradual drift; dual learning leverages the duality of the retrieval module and the prediction module to alleviate the above two problems. Although these methods vary in props and cons, in fact, they could be implemented in a complementary way. Besides, presently, considering the language models like BERT \cite{devlin2018bert}, T5 \cite{raffel2019exploring}, GPT3 \cite{floridi2020gpt}, and etc, have achieved many state-of-the-art results in various applications, it is interesting to combine them into semi-supervised learning in the future.

\bibliographystyle{coling}
\bibliography{umd_scholarly_paper}

\end{document}